\documentclass[10pt]{article}
\usepackage{sdss2020} 
\usepackage{url}
\usepackage{latexsym}
\usepackage{amsmath, amsthm, amsfonts}
\usepackage{algorithm, algorithmic}  
\usepackage{graphicx}
\DeclareMathOperator{\maxarg}{argmax}

\title{Unbiased Estimations based on Binary Classifiers: \\
A Maximum Likelihood Approach}

\author{
  Marco J.H. Puts \\
  Statistics Netherlands / Heerlen, NL \\
  Radboud University / Nijmegen, NL \\
  {\tt mputs@acm.org} \\\And
  Piet J.H. Daas \\
  Statistics Netherlands / Heerlen, NL \\
  Eindhoven Univ. of Technology \\ 
  / Eindhoven, NL \\
  {\tt p.j.h.daas@tue.nl} \\
 }
\date{}
\begin{document}
\maketitle
\begin{abstract}
Binary classifiers trained on a certain proportion of positive items introduce a bias when applied to data sets with different proportions of positive items. Most solutions for dealing with this issue assume that some information on the latter distribution is known. However, this is not always the case, certainly when this proportion is the target variable. 
In this paper a maximum likelihood estimator for the true proportion of positives in data sets is suggested and tested on synthetic and real world data. 
\end{abstract}


\section{Introduction}
An important property of methods applied in (official) statistics is that they provide unbiased estimators. However, this is not always the case for classification methods, especially when one wants to determine the (development of the) proportion of positives in an unknown population. In such cases a bias can affect the target variable tremendously (Figure 1a). 
Classification models are especially sensitive to bias when the scores on the positives and the scores on the negatives overlap \cite{goos_class_2004}, and there are many cases where one does not find a perfect separation between those two, for example, in Natural Language Processing and Image Processing.

Standard bias correction methods assume that information is available on the bias in the data set, see e.g. \cite
{thai-nghe_learning_nodate} and \cite{guo_class_2008}. In particular case that we have an annotated data set, one has the opportunity to look at the confusion matrix or the ROC curve. 

Here, a method is introduced that, based on unlabeled data, is able to correctly estimate the proportion of positives by fitting the distributions of positive and negative items to the probability scores provided by the model. It does this, even when the distributions of the scores of positives and negatives overlap. First, the method is explained, after which it is applied to two data sets.

\section{Methods}

Suppose we have a binary classifier ${B}: {\cal X} \rightarrow (0,1)$, where $\cal X$ is the domain of all feature vectors and the models output is the probability of being positive for a case with features $\overline x$ ($\overline x \in \cal X$). This could, for instance, be a (calibrated) logistic regression model. 
For a training set $\mathbf T$, we can describe this probability by applying Bayes' theorem:
\begin{equation}\label{eq:basic_equality}
{B}(\overline x) = P(+|\overline x,\mathbf T) = \frac { P(\overline x|+, \mathbf T) P(+|\mathbf T) }{P(\overline x|\mathbf T)}.
\end{equation}

Note that \emph{all} probabilities are conditional on the training set $\mathbf T$. When $\mathbf T$ is representative for the population, $P(\overline x|+,{\mathbf T})$ should be representative; i.e. the model should have learned the right features. However, we do not know if the proportion of positive items $P(+|\mathbf T)$ is representative.
Hence, the question is how to get a good estimate for the proportion of positive items.

Let the unknown proportion positives in the data be denoted by $\pi$, and let us define the probability over $b = B(\overline x)$ instead of over $\overline x$. The probability of a score $b$, given the proportion $\pi$ is equal to:
\begin{equation}\label{eq:b|pi}
 P(b|\pi, \mathbf T) =  \pi P(b | +,\mathbf T)  + (1-\pi)P(b | -,\mathbf T).
\end{equation}
Bayes' law gives us the probability of $\pi$:
\begin{equation}
P(\pi|b, \mathbf T)  = \frac {P(b|\pi,\mathbf T) P(\pi)}{P(b|\mathbf T)}.
\end{equation}

Assuming $P(\pi)$ is uniform and $P(b| \mathbf T) = \int_0^1 P(b|\pi,\mathbf T) P(\pi) d\pi$ is a normalization constant, this can be formulated as a likelihood function, which, over the complete data set (${\mathbf B} \subset \{b(\overline x) | \overline x \in \cal X \}$), is equal to:
\begin{equation}
{\cal L}(\pi|\mathbf B,\mathbf T) = \prod_{b \in \mathbf B} P(b|\pi, \mathbf T).
\end{equation}
We can find the maximum likelihood estimate for $\pi$:
\begin{equation}\label{eq:maxlik}
    \hat \pi = \maxarg_\pi \cal L(\pi|\mathbf B, \mathbf T). 
\end{equation}

Since $P(b|+)$ and $P(b|-)$ are independent of the proportion positives in the training set ($\pi_{\mathbf T}$), the maximum likelihood estimate (\ref{eq:maxlik}) is as an unbiased estimator. To approximate $P(b|+)$ and $P(b|-)$ the normalized histograms with $N$ bins for the scores of the positives and the negatives were used. A grid search showed that $N=3$ was optimal for the banknote data (section 3.2). This was used in all examples.

The log likelihood of $\pi$ was also discretized and the maximum likelihood was determined by finding the index with the highest log likelihood.

\section{Results}

\subsection{Simulated data}
The first data set consisted of simulated data from two normal distributions, separated by two standard deviations. From both distributions, 500 values were randomly drawn, with a proportion of $p$ from the positive distribution and with a proportion $1-p$ from the negative distribution.

\begin{figure}
\centering
\makebox{\includegraphics[scale=0.45]{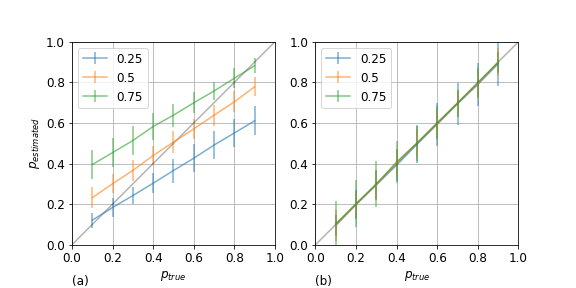}}
\caption{Estimates of the proportion positives in simulated evaluation data ($y$) for models trained on 25\% (blue), 50\% (orange) and 75\% (green) positives against their true proportion ($x$). Panel (a) shows the results of each model. Panel (b) shows the maximum likelihood estimates of each model. }
\label{betasim}
\end{figure}
For three different proportions of positives ($p \in \{0.25, 0.5, 0.75\}$), data was generated on which a logistic regression model was trained. The models were used to predict the labels of 9 evaluation sets with a positive proportion varying from $0.1$ to $0.9$ (step $0.1$). The whole procedure was repeated 100 times. In Figure 1a, the estimated proportion of positive items by each model ($y$) for each evaluation set is shown against the real proportion of positive items included ($x$). The error bars indicate the 95\% interval of the estimates. Ideally, the estimated proportion of positive items should be identical to the real proportion of positive items (gray line). Evidently, this is not the case. In Figure 1b, the maximum likelihood estimated proportions of positive items (see Eq. \ref{eq:maxlik}) are plotted for each model. As can be seen, the estimations are all centered around the true proportions. 

\subsection{Banknote Authentication data}
As a second data set, the banknote authentication data set (see https://bit.ly/35K3m3i) was used. This data set contains features measured on photographs of authentic and counterfeit banknotes. Four features are included: variance, skewness and curtosis of the wavelet transformed image and the entropy of the image. For each banknote, its authenticity (True/False) is also given. 

Fitting a logistic regression model, to estimate the authenticity, on all 4 features resulted in a perfect fit. For that reason, only the skewness and curtosis features were used. A random selection of 411 items (195 pos. and 216 neg.) were used for the training set. Figure 2 shows the results of the model on 9 banknote evaluation sets with a positive proportion of $0.1$ to $0.9$ (step $0.1$) repeated a 100 times. The proportion of positives estimated by the model shows strong biases (Figure 2a), the maximum likelihood based estimates do not (Figure 2b).
\begin{figure}
\centering
\makebox{\includegraphics[scale=0.45]{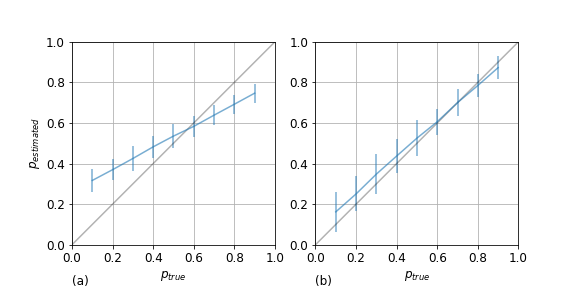}}
\caption{Estimates of the proportion positives in banknote data ($y$) for the two feature trained model against the true proportions included ($x$). Panel (a) shows the results of the model. Panel (b) shows the maximum likelihood estimates. }
\label{betabank}
\end{figure}
\section{Discussion}
The results show that the maximum likelihood method developed gives good, unbiased estimates for the proportion positives in binary classification problems. It does this by fitting the distributions $P(b|+)$ and $P(b|-)$ 'learned' to the distribution of scores obtained in the -to be- classified data set.
In this paper, distributions were approximated by using the normalized histograms of the positives and negatives scores. A grid search on the number of bins revealed that the model worked best with three bins. Using parametric distributions could be a way to improve it.
Furthermore, it should be possible to extend this approach to multi class classifications. Since this also means that a maximum needs to be found in a multidimensional likelihood space, other methods, such as Markov Chain Monte Carlo sampling, may need to be incorporated.

\bibliographystyle{sdss2020} 
\bibliography{references.bib}
\end{document}